\documentclass[preprint,12pt, a4paper]{elsarticle}

\usepackage{amssymb}
\usepackage{hyperref}
\usepackage{ulem}
\DeclareFontFamily{U}{euc}{}
\DeclareFontShape{U}{euc}{m}{n}{<-6>eurm5<6-8>eurm7<8->eurm10}{}%
\DeclareSymbolFont{AMSc}{U}{euc}{m}{n} 
\DeclareMathSymbol{\umu}{\mathord}{AMSc}{"16} 
\usepackage{glossaries}
\newacronym{uct}{$\umu$-CT}{micro-Computed Tomography}

\usepackage{lineno}

\journal{SoftwareX}

\begin{document}

\begin{frontmatter}



\title{TomoSAM: a 3D Slicer extension using SAM for tomography segmentation}


\author[AMA]{Federico Semeraro\corref{cor1}\fnref{label2}}
\ead{federico.semeraro@nasa.gov}
\author[AMA]{Alexandre Quintart}
\author[AMA]{Sergio Fraile Izquierdo}
\author[Stanford]{Joseph C. Ferguson}

\address[AMA]{AMA Inc. at NASA Ames Research Center, Moffett Field, CA 94035}

\address[Stanford]{Dept. of Mechanical Engineering, Stanford University, Stanford, CA 94305}

\begin{abstract}

TomoSAM has been developed to integrate the cutting-edge Segment Anything Model (SAM) into 3D Slicer, a highly capable software platform used for 3D image processing and visualization. SAM is a promptable deep learning model that is able to identify objects and create image masks in a zero-shot manner, based only on a few user clicks. The synergy between these tools aids in the segmentation of complex 3D datasets from tomography or other imaging techniques, which would otherwise require a laborious manual segmentation process. The source code associated with this article can be found at \href{https://github.com/fsemerar/SlicerTomoSAM}{https://github.com/fsemerar/SlicerTomoSAM}. 

\end{abstract}

\begin{keyword}
3D Segmentation \sep SAM \sep Slicer \sep Tomography
\end{keyword}

\end{frontmatter}

\section{Introduction}\label{sec:intro}

The quantitative analysis of 3D imaging data, either for medical or material science purposes, typically requires the segmentation of the different components present within the image - often a challenging and time-consuming process. For this reason, many deep learning techniques have been proposed to perform the segmentation task, but they usually require a large amount of labeled data to train, as well as considerable computational resources. 

Our work is motivated by the objective of modeling Thermal Protection Systems (TPS), in order to estimate their material properties and predict their response to the harsh environmental conditions experienced during atmospheric entry. The Porous Microstructure Analysis (PuMA) software \cite{ferguson2021update} was developed to provide a robust and efficient framework for the digital characterization of 3D microstructures. These digital models are usually obtained through micro-Computed Tomography (CT), an imaging technique that can resolve the structure of a material at a sub-micron scale in 3D, and even temporally in 4D. Over the past decade, this method has revolutionized the materials science field for its ability to non-destructively analyze a material's microstructure. PuMA provides the capability of computing a comprehensive spectrum of properties, from morphological (e.g. specific surface area, volume fractions, mean intercept length, orientation \cite{semeraro2020anisotropic}) to physical (e.g. conductivity \cite{semeraro2021anisotropic}, elasticity \cite{fraile2022multi}, permeability \cite{lopes2023simulation}, tortuosity \cite{ferguson2022continuum}) and chemical (e.g. oxidation \cite{ferguson2016modeling}). 

\begin{figure*}[!htb]
\centerline{\includegraphics[width=1.01\linewidth]{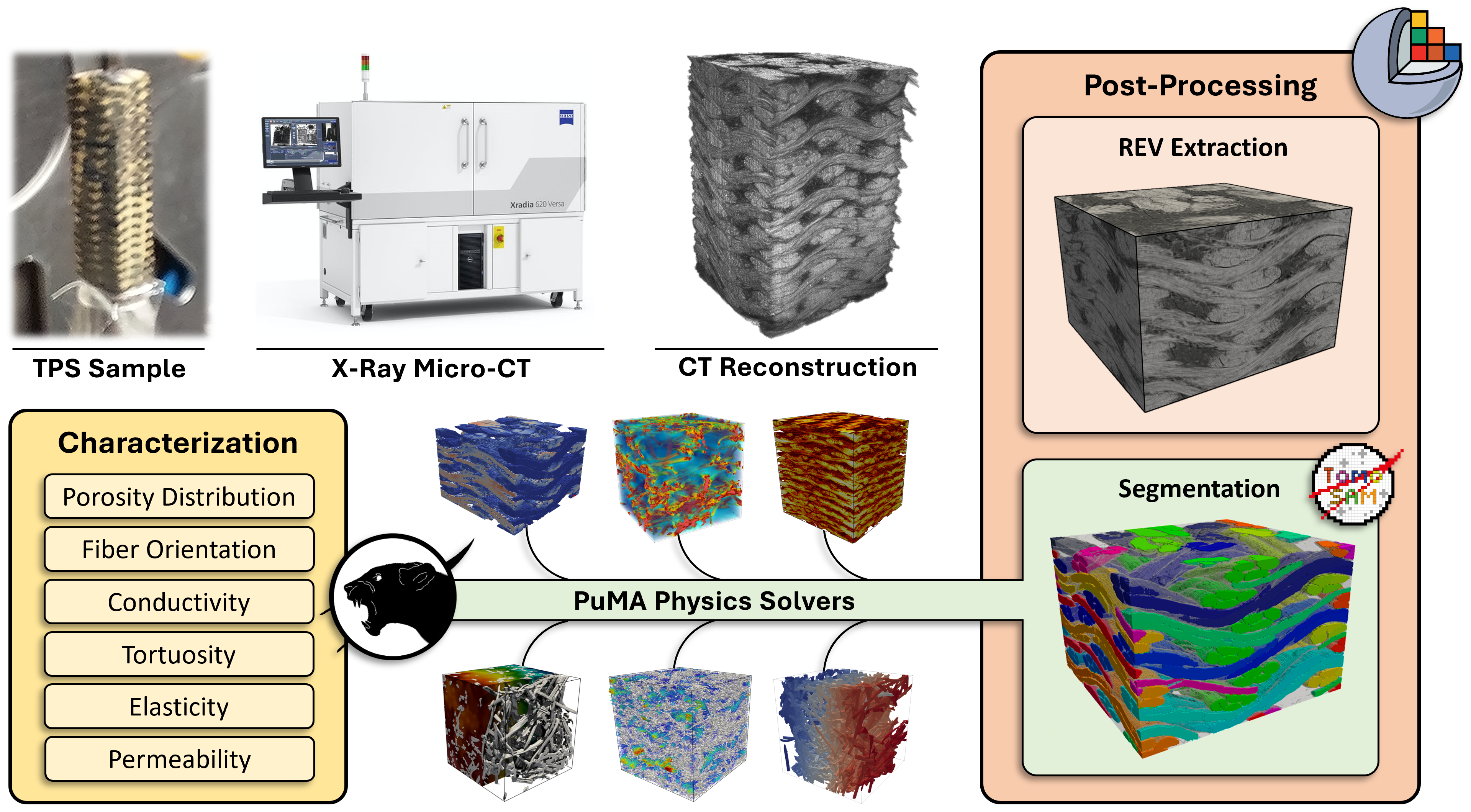}}
\caption{Workflow for the digital characterization of a sample using micro-CT data.}
\label{fig:ct_workflow}
\end{figure*}

A typical workflow to analyze a new sample using micro-CT is shown in Figure \ref{fig:ct_workflow}, where a composite carbon fiber weave is being considered. It starts from the preparation of the sample, followed by the X-ray scanning, tomographic reconstruction, and image post-processing of the digital model to capture a Representative Elementary Volume (REV). Before computing the effective material properties using software like PuMA, the computational domain must be first segmented into its different components. For some CT scans, simple thresholding based on X-ray attenuation can be used for segmentation. Many cases, however, require more advanced segmentation methods, using software like 3D Slicer \cite{pieper20043d}, Dragonfly \cite{makovetsky2018dragonfly}, or Fiji \cite{schindelin2012fiji}. Even with these tools, segmentation is often difficult, labor-intensive, and time-consuming. 

For 3D image segmentation, there are three broad categories in which the subdivision of the components occurs: instance, semantic, and panoptic. In instance segmentation, the goal is to identify and delineate individual objects (``things''), assigning a unique label to each object instance while also providing precise pixel-level (or voxel-level) masks. An example would be to label each tow within a weave with a different ID. Semantic segmentation, on the other hand, focuses on assigning a class label to each pixel in an image, grouping pixels with similar characteristics together. This task aims to provide a holistic understanding of the scene by segmenting it into semantically meaningful regions (``stuff''). Following from our example, this would mean assigning a unique ID to all the tows running in one direction, one for all the tows running in the other direction, one for the resin, and one for the void space. Panoptic segmentation combines elements from both instance and semantic segmentation, aiming to produce a comprehensive and unified output. It involves not only labeling each pixel with a class but also identifying and separating individual instances within each class. Our approach in this work has the objective of targeting any of these segmentation tasks through a versatile, promptable tool that can adapt to many 3D data sources.

New developments in the computer vision and deep learning fields have significantly streamlined the segmentation process of 2D images. More specifically, Vision Transformers (ViTs) \cite{dosovitskiy2020image} have garnered significant attention for their ability to achieve state-of-the-art performance on various visual recognition tasks. Unlike traditional Convolutional Neural Networks (CNNs), which rely on a hierarchical arrangement of convolutional layers, ViTs adopt a self-attention mechanism to capture long-range dependencies in the input image. This mechanism allows the model to efficiently track global interactions among different regions, enabling the extraction of meaningful visual features. Concurrently, Masked Autoencoders (MAE) \cite{he2022masked} have become popular for their ability to efficiently learn representations in a self-supervised manner by randomly masking out a portion of the input image and then training to reconstruct the missing pixels. This process, which can be run on a large dataset of unlabeled images, forces the model to learn the underlying structure of the image in a latent space, which can then be used for downstream tasks such as classification or object detection. The recently published Segment Anything Model (SAM) \cite{kirillov2023segment} uses a combination of these techniques, namely a computationally-heavy MAE-ViT as its image encoder, a flexible prompt encoder using positional encodings, and a significantly lighter Transformer block as its mask decoder. This architectural choice is crucial because it enables the creation of the necessary embeddings on a different machine with access to a CUDA-compatible GPU, and then run a real-time interactive mask prediction on a normal laptop. SAM was trained on more than 1 billion masks over 11 million images, enabling it to create a generalized and transferable representation that works effectively on unseen data.

In this work, we introduce TomoSAM, a module of the PuMA software that simplifies and accelerates complex segmentation tasks. TomoSAM integrates SAM into the widely used 3D Slicer software, utilizing the capabilities of this advanced deep learning architecture to detect objects through a zero-shot transfer approach with only a small number of user-provided points. In addition, we demonstrate that our approach generalizes well to a variety of 3D voxel data, by using it in conjunction with some of Slicer's existing functionality. While TomoSAM was developed with multidisciplinary applications in mind, it is not intended for clinical use.

\section{Methods}\label{sec:methods}

\begin{figure*}[!htb]
\centerline{\includegraphics[width=\linewidth]{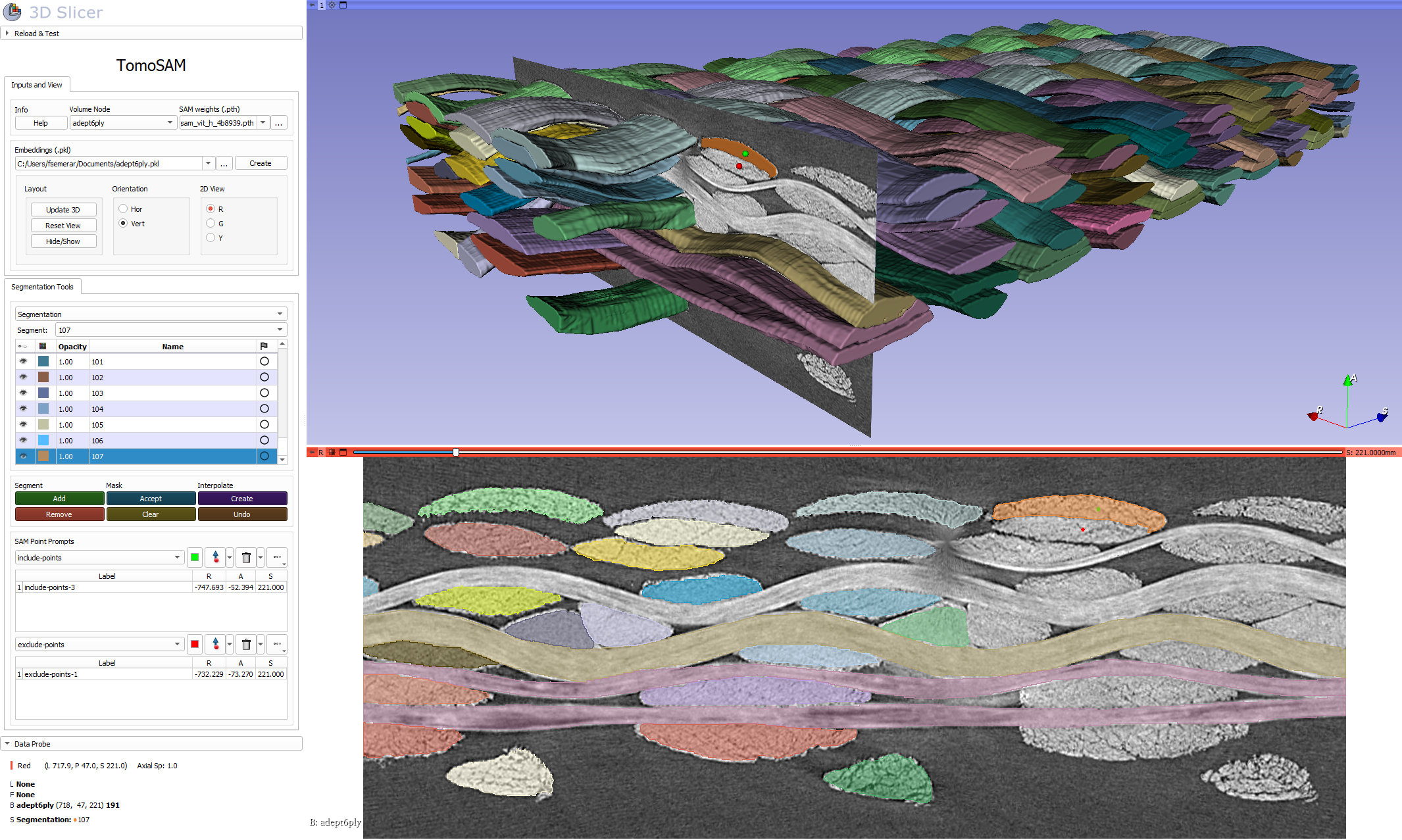}}
\caption{Segmentation of a tomography dataset of a carbon fiber weave using TomoSAM.}
\label{fig:tomosam_weave}
\end{figure*}

\begin{figure*}[!htb]
\centerline{\includegraphics[width=\linewidth]{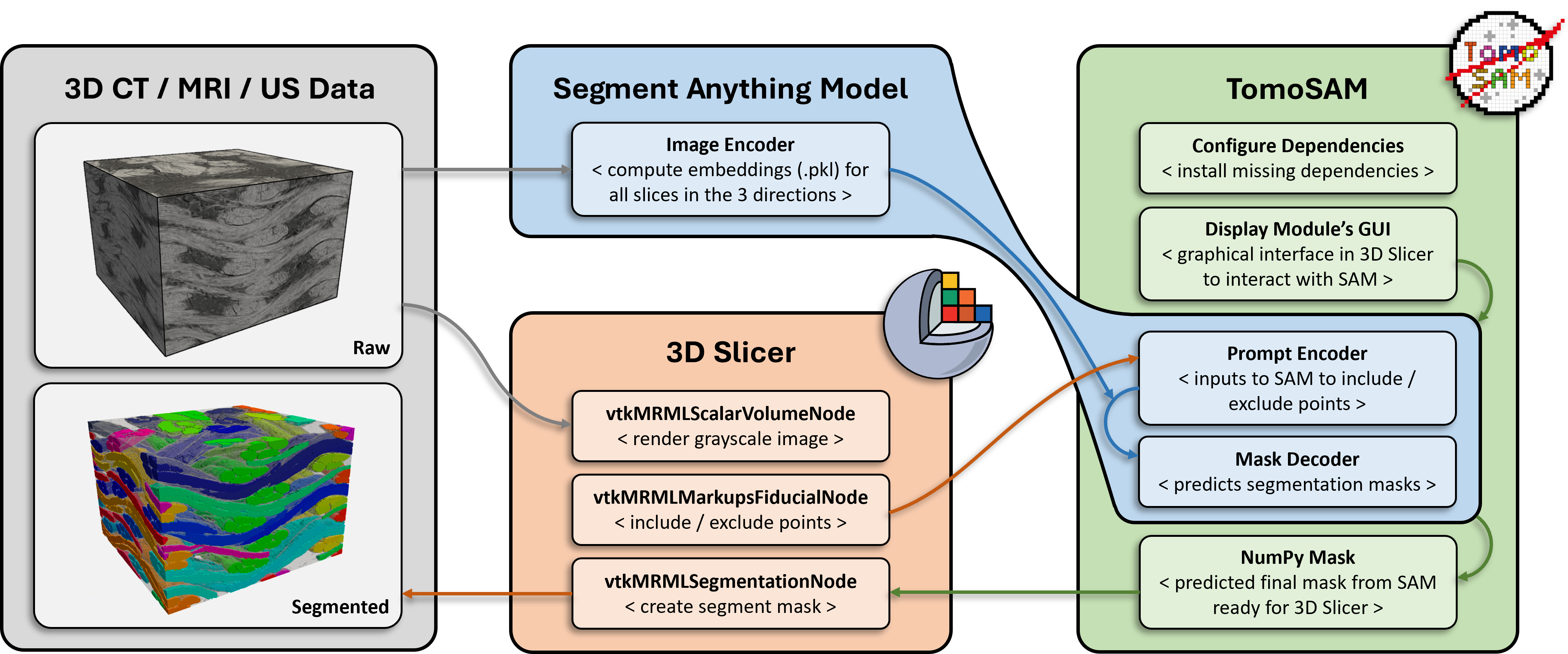}}
\caption{TomoSAM's usage, architecture, and communication system.}
\label{fig:tomosam_diagram}
\end{figure*}

The TomoSAM extension is shown in action in Figure \ref{fig:tomosam_weave} to segment a carbon fiber weave, and its architecture is displayed in Figure \ref{fig:tomosam_diagram}. Before the interactive segmentation in 3D Slicer can begin, a pre-processing step is required to generate the image embeddings by slicing along the three Cartesian directions and iteratively calling the \texttt{set\_image} method from the \texttt{SamPredictor} class, which generates the feature embeddings through SAM's image encoder. This can be performed through the provided Jupyter notebook, which can either be run online on Google Colaboratory or locally by installing the Anaconda environment. As previously mentioned, this step is computationally intensive and it should be therefore run on a machine with a CUDA-compatible GPU. The separation between the embeddings computation and the promptable segmentation is important to optimally leverage SAM's architecture and make the overall process more interactive and user-friendly \cite{liu2023samm}. It is important to note that applying contrast-enhancement image filters before the creation of the embeddings was observed to significantly help SAM discern between object boundaries.

Once an embeddings file (.pkl) has been generated, the 3D image along with its embeddings can be imported into TomoSAM within a Slicer instance, which can be run on a simple CPU machine. The image import automatically generates three node types: a \textit{volume} node, a \textit{segmentation node}, and two \textit{markup} nodes. The \textit{volume} node contains the raw 3D grayscale image. This is only used for visualization purposes, so that the user can choose points correctly; the image is not used by SAM's Predictor, only its pre-computed embeddings are queried. The \textit{segmentation} node holds the masks and segments of the different sample components, which are populated during the segmentation process. Finally, the two \textit{markup} nodes are used to store the \textit{include} and \textit{exclude} points that the user can add interactively. These prompted points are passed to the SAM predictor, which produces their positional encoders. The mask decoder translates the image embedding, prompt embeddings, and an output token into a mask, which is sent back to 3D Slicer for visualization. Note that, thanks to SAM's architectural choices, these steps can run with minimal latency even for domains with $1e7-1e9$ voxels (e.g. 0.5\,--\,10 seconds on a 8-core laptop without access to a dedicated GPU).

After the addition of a few masks, e.g. one every 20-30 voxels, Slicer's ``fill between slices'' function \cite{albu2008morphology}, which is integrated within TomoSAM, can be used to interpolate between the missing frames and complete the 3D segment.

% \section{Results}\label{sec:results}

% datasets: 
% - brain
% - spleen (http://medicaldecathlon.com/)
% - lung (https://github.com/rbumm/SlicerLungCTAnalyzer)
% - liver (LiTS https://competitions.codalab.org/competitions/17094)
% - 4 adept plies

\section{Conclusion}\label{sec:conclusion}

In this paper we have presented TomoSAM, a novel 3D segmentation tool that harnesses the power of the Segment Anything Model. The TomoSAM Slicer extension addresses the challenges associated with accurate and efficient segmentation of 3D images, offering significant advancements in the computer vision field, as well as across various domains including medical imaging and materials science. Through comprehensive experimentation and evaluation, SAM was observed to outperform existing segmentation methods in terms of accuracy, speed, and robustness. By leveraging its ability to handle complex structures and adapt to different image modalities, this tool has the potential to revolutionize the analysis and interpretation of 3D imaging data. Furthermore, TomoSAM's user-friendly interface and streamlined workflow, which separates the computationally-heavy embeddings creation and assisted manual segmentation, make it accessible to a wide range of researchers and engineers. Its automated segmentation capabilities not only save valuable time and resources but also provide reliable and consistent results, reducing human bias and subjectivity.

It is important to note that the development of this software is an ongoing process, and there is room for further refinement and optimization. Future research will focus on addressing potential limitations, such as handling noisy or low-contrast datasets. Overall, the introduction of the Segment Anything Model in the context of segmenting 3D voxel data represents a significant advancement in the field, with the potential to revolutionize how tomographic data is analyzed and interpreted.

\section*{Acknowledgements}
This work was supported by the NASA Entry System Modeling project (J. Haskins, Project Manager; A. Brandis, Principal Investigator), by the NASA Space Technology Research Fellowship program (Award No. 80NSSC19K1134), and by the NASA contract NNA15BB15C to Analytical Mechanical Associates, Inc. We would like to thank J. Thornton and J. Meurisse for the review of the manuscript, as well as P. Wercinski, A. Cassell, F. Panerai, and K. Hendrickson for providing the woven data. Finally, we would like to acknowledge A. Borner, L.J. Abbott, M. Stackpoole, K. Doss, and K.R. Wheeler for useful discussions.

% \begin{table*}[]
% \centering
% \begin{tabular}{|l|p{6.5cm}|p{6.5cm}|}
% \hline
% \textbf{Nr.} & \textbf{Code metadata description} & \textbf{Please fill in this column} \\
% \hline
% C1 & Current code version & 3.0 \\
% \hline
% C2 & Permanent link to code/repository used for this code version & \href{https://github.com/fsemerar/SlicerTomoSAM}{https://github.com/fsemerar/SlicerTomoSAM} \\
% \hline
% C3 & Code Ocean compute capsule & N/A\\
% \hline
% C4 & Legal Code License   & NASA Open Source License v1.3 \\
% \hline
% C5 & Code versioning system used & git \\
% \hline
% C6 & Software code languages, tools, and services used & Python, PyTorch, OpenCV \\
% \hline
% C7 & Compilation requirements, operating environments \& dependencies & Linux, MacOS, Windows\\
% \hline
% C8 & If available Link to developer documentation/manual & \href{https://github.com/fsemerar/SlicerTomoSAM} \\
% \hline
% C9 & Support email for questions & \href{mailto:federico.semeraro@nasa.gov}{federico.semeraro@nasa.gov}\\
% \hline
% \end{tabular}
% \caption{Code metadata.}
% \label{} 
% \end{table*}

% \newpage
% \footnotesize
\bibliographystyle{elsarticle-num} 
\bibliography{references}

\end{document}